\documentclass[lettersize,journal, twoside]{IEEEtran}
\usepackage{amsmath,amsfonts}
\usepackage{algorithmic}
\usepackage{algorithm}
\usepackage{array}
\usepackage{textcomp}
\usepackage{stfloats}
\usepackage{url}
\usepackage{verbatim}
\usepackage{cite}
\usepackage{graphicx}

\usepackage{graphics}
\usepackage{epsfig} % for postscript graphics files
\usepackage{subcaption}
\usepackage{setspace}
\usepackage{bm}
\usepackage{upgreek}
\usepackage[labelsep=period]{caption} 
\usepackage[british]{babel} 

\begin{document}

\title{Gaussian Process Mapping of Uncertain Building Models with GMM as Prior}

\author{Qianqian Zou, Claus Brenner and Monika Sester
        % <-this % stops a space
\thanks{Manuscript received: March 3, 2023; Revised:
June 11, 2023; Accepted: July 15, 2023. 
This paper was recommended for publication by Editor Javier Civera upon evaluation of the Associate Editor and Reviewers’ comments.
This work was supported by the German Research Foundation (DFG) as part of the Research Training Group i.c.sens. (corresponding author: Qianqian Zou)}% <-this % stops a space
\thanks{The authors are with the Institut of Cartography and Geo-informatics in Leibniz University Hannover, Appelstraße 9a 30167 Hannover, Germany (email: qianqian.zou@ikg.uni-hannover.de; claus.brenner@ikg.uni-hannover.de; monika.sester@ikg.uni-hannover.de).}
\thanks{Digital Object Identifier (DOI): 10.1109/LRA.2023.3303694.}
}

% The paper headers
% \markboth{IEEE ROBOTICS AND AUTOMATION LETTERS. PREPRINT VERSION. ACCEPTED AUGUST, 2023}
% {Shell \MakeLowercase{\textit{ZOU et al.}}: A Sample Article Using IEEEtran.cls for IEEE Journals}
\markboth{IEEE ROBOTICS AND AUTOMATION LETTERS. PREPRINT VERSION. ACCEPTED AUGUST, 2023}
{ZOU \MakeLowercase{\textit{et al.}}: GP Mapping of Uncertain Building Models with GMM as Prior}

% \IEEEpubid{0000--0000/00\$00.00~\copyright~2021 IEEE}
% Remember, if you use this you must call \IEEEpubidadjcol in the second
% column for its text to clear the IEEEpubid mark.

\maketitle

\begin{abstract}
Mapping with uncertainty representation is required in many research domains, especially for localization. Although there are many investigations regarding the uncertainty of the pose estimation of an ego-robot with map information, the quality of the reference maps is often neglected. To avoid potential problems caused by the errors of maps and a lack of uncertainty quantification, an adequate uncertainty measure for the maps is required. In this letter, uncertain building models with abstract map surfaces using Gaussian Processes (GPs) are proposed to describe the map uncertainty in a probabilistic way. To reduce the redundant computation for simple planar objects, extracted facets from a Gaussian Mixture Model (GMM) are combined with an implicit GP map, also employing local GP-block techniques. The proposed method is evaluated on LiDAR point clouds of city buildings collected by a mobile mapping system. Compared to the performance of other methods such as OctoMap, GP Occupancy Map (GPOM), Bayesian Generalized Kernel OctoMap (BGKOctoMap), Local automatic relevance determination Hilbert map (LARD-HM) and Gaussian Implicit Surface map (GPIS), our method achieves a higher Precision-Recall AUC for the evaluated buildings.
\end{abstract}

\begin{IEEEkeywords}Mapping, Probability and statistical methods, Laser-based, Uncertainty representation.
\end{IEEEkeywords}

\section{Introduction}
\IEEEPARstart{M}{aps} play a crucial role in assisting autonomous systems, such as robots or automated vehicles, to comprehend unknown and complex scenes, enabling them to locate and navigate themselves accurately and safely. However, the uncertainties associated with maps significantly impact the performance of localization, navigation security, and collision avoidance of the autonomous system. With laser scanners, the environment can be represented in a straightforward way with point clouds, and further modelled using structured maps or abstract surface models \cite{abstract_map}. The common maps are, e.g., high-definition (HD) vector maps \cite{hd-map_localization}, acquired with high geometric and semantic accuracy, 2D or 3D-grids with occupancy probabilities, such as OctoMaps \cite{octomap}, or 3D-City models with various levels of detail. Uncertainty sources in the environment map are inevitable due to, e.g., ambiguous environment information, sparsity, occlusions and noise in the measurements \cite{review}. Hence, when the map is treated as an ideal perfect reference in localization and path-planning tasks without considering its quality, this can result in severe errors, e.g., the wrongly modelled occluded buildings might cause a bias in the estimated pose. Additionally, one may use multiple sensors to capture the environment or may have the demand of updating the existing maps with new data. In those scenarios, a good uncertainty measure of the maps created by different sensors and agents is essential for data fusion and motion planning purposes \cite{slam-uncertainty}. Although many existing studies about uncertainty models are found in automated driving related fields, they are mostly for pose estimation. There is a lack of sufficient research describing the accuracy and precision of maps.

In the field of simultaneous localization and mapping (SLAM), uncertainty-aware tasks require a reasonable expression of map uncertainty. There are some existing methods estimating the uncertainty of  maps to improve localization accuracy or make the results more robust\cite{ndt,interval_localization, abstract_map}. For example, Biber and  Straßer \cite{ndt} proposed to represent the environment using normal distribution transform (NDT) maps, where the map uncertainty is represented by a grid of distributions. However, it assumes independent discontinuous distributions of the neighboring cells and it has been found to result in a higher uncertainty at cell boundaries \cite{gmm3}. Javanmardi et al. \cite{abstract_map} introduced the idea to build abstract vectors and planar surface maps of buildings and ground, where the uncertainty is given by the normal distribution generated from vectors and planes. Building maps with fixed interval uncertainties have been used in hybrid interval-probabilistic localization \cite{interval_localization}. Occupancy maps with uncertain occupied cells are also widely used in localization, whose uncertainties are represented by the occupancy probabilities. 

All these existing studies prove the importance of quantifying the uncertainty of maps. However, it remains challenging to map the continuous space from noisy sparse point clouds and assign reasonable uncertainties to it. There is still a research gap in measuring the spatial uncertainty of maps in urban areas. There are many man-made structures with relatively regular geometrical shapes, where buildings are often the most important ones for localization, e.g. as used in \cite{abstract_map, interval_localization}. Thus, a detailed spatial uncertainty representation of building models in urban scenes is of great interest to explore.

The choice of an adequate uncertainty measure in environment mapping is critical.  To tackle this problem, the uncertainty and completeness of a map can be described in a probabilistic fashion. GPs with their data-driven and probabilistic nature have proven to be a powerful tool for the quantification of uncertainty in various research fields. Recently, GPIS maps \cite{gpis1, gpis2} have been proposed for continuous environmental mapping. The posterior mean and covariance kernel of a GP serve as the best estimation of the surface distance field and the uncertainty measure, respectively. GP Occupancy Mapping (GPOM)\cite{gpom} demonstrates how to predict a continuous occupancy map from sparse noisy point clouds with GP inference. The occupancy probability of unexplored space with no rays passing through or hitting is regressed by spatial neighbors in a GP, including well-calibrated uncertainties. It provides the possibility to capture the underlying correlation between spatially neighboring cells. The continuous inference gives the generalizability to building a map of desired resolution for diverse applications. However, the drawback of GP lies in the high computational cost due to matrix inversion operations, and this drawback limits the applicability of GPs for large datasets. In order to achieve an acceptable computational efficiency, many approximation approaches such as the partitioning of the spatial world into subsets, the fusion of local GPs with Bayesian Committee Machines (BCMs) or a sparse approximation of GPs have been proposed \cite{gpom, gpmap, gpom_fast,sparsekernel,multi-sparse-gp}.  To avoid the unreliable prediction from successive BCM updates, the Bayesian generalized kernel OctoMap (BGKOctoMap) \cite{bgk} leverages Bayesian kernel inference and sparse kernels to perform a stable inference-based occupancy mapping. Jadidi et al.  \cite{gp_frontier} expanded the occupancy mapping to probabilistic geometric frontiers computed efficiently using the gradient of the GP occupancy map. 

% *********** introduction to Hilbert map
Hilbert Maps, as another continous mapping technique, have been introduced by Ramos et al.\cite{hm1, hm2} to efficiently map the complex real world by operating on a high-dimensional feature vector. With efficient stochastic gradient descent optimization, it can achieve comparable performance to GP mapping with less time. Inspired by GP, Duong et al.\cite{hm3} introduced a probabilistic formulation that utilizes a sparse set of relevance vectors to model obstacle boundaries.

However, these methods lack proper uncertainty quantification for their mapping. The uncertainty inherent in a GP map has not been sufficiently investigated, which can be exploited to distinguish unknown areas without enough information. This highlights the need to evaluate the reliability of the inferred occupancy probabilities. This issue has been addressed in recent work by Pearson \cite{uncertain_bgk}. Additionally, occasional discontinuities in the environment can pose challenges for GP-based mapping, as GPs are often used for continuous targets. This effect is not well-studied in the current literature on GP mapping either.

Another group of research is learning structural surface models from measurements to probabilistically represent the perceived space. Thrun et al. \cite{{multiplane}} fitted a set of rectangular flat surfaces to compose three-dimensional (3D) maps, with a group of parameters optimized by Expectation Maximization (EM) and the number of planes estimated by a Bayesian prior. With similar spirit but higher fidelity in approximating diverse arbitrary environments, Gaussian Mixture Models (GMMs) are employed as a semi-parametric tool with large numbers of components to extract the planar models \cite{gmm1} or obtain a high-fidelity representation of sensor observations\cite{gmm2, gmm3, gmm4}, estimating Gaussian mixtures instead of 3D planes. Nevertheless, the number of components has a great impact on the performance of GMM-based approaches and it is a non-trivial task to choose a proper number.

In this work, GP is used to obtain an inference-based surface model of structured buildings with GMM priors, leveraging the flexibility of non-parametric methods in arbitrary structures as well as the scalability of parametric approaches in large environments. When exploring large urban scenes, buildings are one of the most important structures for localization, which have relatively simple geometry and are easy to extract. GMMs are first applied to extract the main planar surfaces with a few components, while GPs are used subsequently to carve the irregular parts. In GP inference, the sets of local GPs corresponding to certain planar objects are applied to adapt the discontinuities and reduce redundant computation, while the sparse kernel and data block partitioning techniques similar to the prior work \cite{gpmap, gpom_fast} are utilized to further improve the computational efficiency.

The contributions of this work are the following:
\begin{itemize}
\item We map uncertain building models in a probabilistic manner using laser scanning point clouds, which can be applied in various downstream tasks and can be updated probabilistically. This method addresses the issue of a lack of uncertainty quantification for reference maps. The merit of this uncertainty representation is revealed in the experiments where better results are achieved when using accurate inference.

\item GMM planes and local GP blocks are combined to yield a map with efficient surface prediction and to maintain a faithful uncertainty description.
Both GMM planes and GP functions represent the statistical uncertainty of the mapping process.  

\item We evaluate the mapping performance and the faithfulness of the uncertainty description with real-world LiDAR point clouds, collected by a mobile mapping system (MMS). The Precision-Recall (PR) curve and the area under the curve (AUC) are used as metrics to evaluate the results of OctoMap, fast GPOM, BGKOctoMap, LARD-HM, GPIS, and our method. A comparison to the receiver operating characteristic (ROC) AUC is also discussed. 
\end{itemize}

\begin{figure*}[thpb]
      \centering
      \includegraphics[width=0.7\linewidth, trim={0.5cm 0cm 0.5cm 0.5cm},clip]{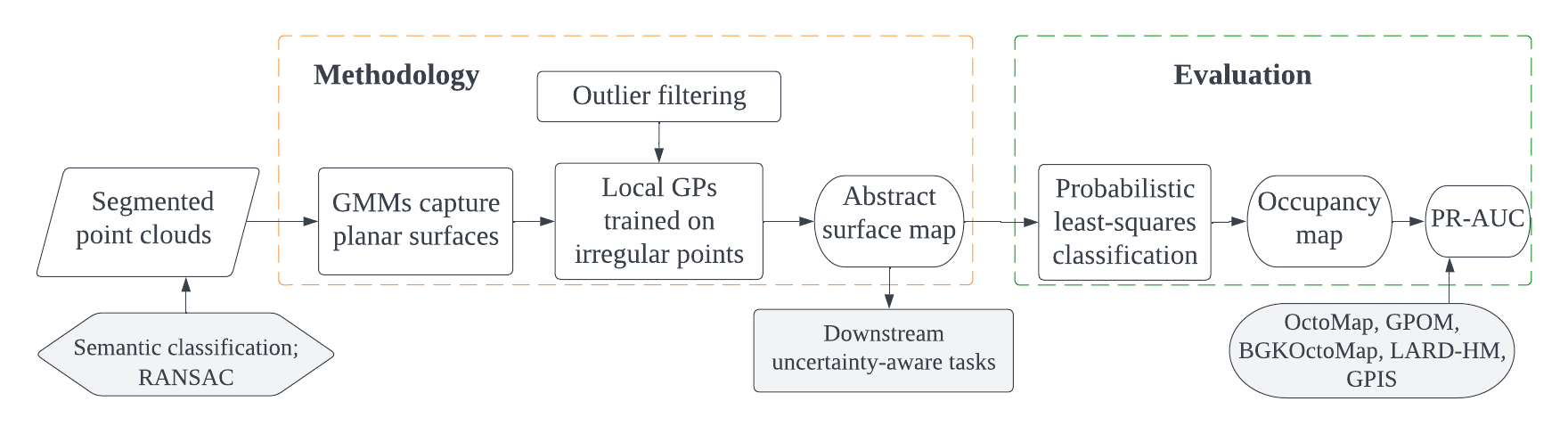}
      \caption{Overall workflow.}
      \label{fig:flowchart}
 \end{figure*}

The rest of the letter is structured as follows. Section II defines the problem we aim to solve as well as the representation of building maps with uncertainty measures. In Section III, a brief introduction to GMMs and GPs is presented. We explain how GPs are used to model non-planar surfaces with a GMM prior, representing the uncertain building façade surface. The strategies for parameterizing the discontinuities, reducing computational efforts, and outlier filtering are explained in detail. In Section IV, experiments using real LiDAR data collected by MMS are presented, and qualitative results as well as a quantitative evaluation with state-of-the-art benchmarks are shown. Finally, we provide our conclusions and an outlook for future work in Section~V. Figure \ref{fig:flowchart} illustrates the overall workflow of the mapping process and evaluation.

\section{Problem Statement}

%  Concerning the problem of mapping uncertain building models as the input map for an automotive robot in outdoor scenarios, we aim at offering a quantitative uncertainty measure for building models,
The objective is to produce maps for automotive robots in outdoor scenarios, which consist of building models including quantitative uncertainty measures for each surface point, as illustrated in Figure \ref{fig:b15}. More specifically, the goal of the mapping problem is to estimate continuous façade surface: determining the relative surface depth and its uncertainty at any given query point. %Considering that sensors provide uncertain points with noise and errors, the uncertainty measure should properly describe the quality of the map. 

Data sources are building points obtained from LiDAR range measurements. (Classification of the raw measurements to extract the building points is performed during a preprocessing step.) Due to the measurement process with a MMS, there are invisible areas occluded by other objects (cars, trees) or also self-occlusion. The resulting map should take these factors into account by properly describing the uncertainty measures. Such a representation also allows for integrating data from different sensors of different accuracy.

Although buildings are 3D objects, they can be modelled as being composed of a set of facades, each of which consists of planar protrusions, extrusions, slopes (non-vertical planes) and non-planar shapes, leading to a 2.5D representation.

Each individual façade of a building is then modelled with two parts: 
%In each local façade coordinate, we consider the depth value over the façade plane as a continuous representation of the building surface. It is modelled with two parts: 
(1)~regular planar parts are segmented by a GMM; (2)~deviations from planar segments are modelled by GPs to capture arbitrary shapes. The planar parts can be modelled using plane parameters and boundary points indicating the location and the shape of each local planar facet. These parameters are optimized in GMM to provide a prior surface estimation. In this way, the main façade plane, as well as  protrusions and extrusions can be specified with the corresponding plane layers resulting from a GMM. The remaining non-planar parts are represented as relative surfaces via non-parametric approaches, namely GP in our case, which describes the surface as a function with depth values of each point. Given a point $\boldsymbol{x} \in \mathbb{R}^2$ on the local projected surface plane, the distance $s$ to the prior mean surface is:

\begin{figure}[htpb]
      \centering
      \includegraphics[width=0.48\linewidth]{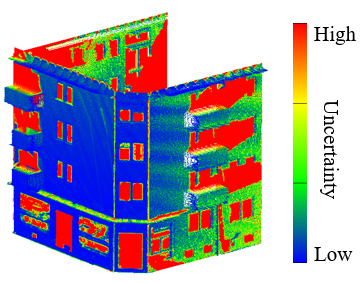}
      \caption{
      Depiction of a building model where for each location an uncertainty can be
      queried (indicated by color).
      }
      \label{fig:b15}
 \end{figure}
 
\begin{equation}
\label{eq:surface}
s = f(\boldsymbol{x}) \left\{ \begin{array}{cl}
= 0 & : \ \text{ on the mean surface}, \\
<0 & : \ \text{ on an extrusion},  \\
>0 & : \ \text{ on a protrusion}.
\end{array} \right.   
\end{equation}

This GP surface representation provides an estimate of depth regarding the local planar segment at any location on the surface.
%as well as the corresponding uncertainty of the estimation. 
%
Combining both components, the surface depth $d(\boldsymbol{x})$ is given by:
\begin{equation}
\label{eq:statement}
d(\boldsymbol{x}) = g(\boldsymbol{x}) + f(\boldsymbol{x}), 
\end{equation}
where $g(\boldsymbol{x})$ is the expected plane depth from the GMM and $f(\boldsymbol{x})$  denotes the local surface distance from Equation \eqref{eq:surface}.
%with regard to the planar component.

\section{Uncertain Building Surface Model}
\begin{figure*}[ht!]
\begin{subfigure}[t]{.33\textwidth}
  \centering
  % include second image
  \includegraphics[width=0.73\linewidth, trim={3cm 12.8cm 0cm 15.5cm},clip]{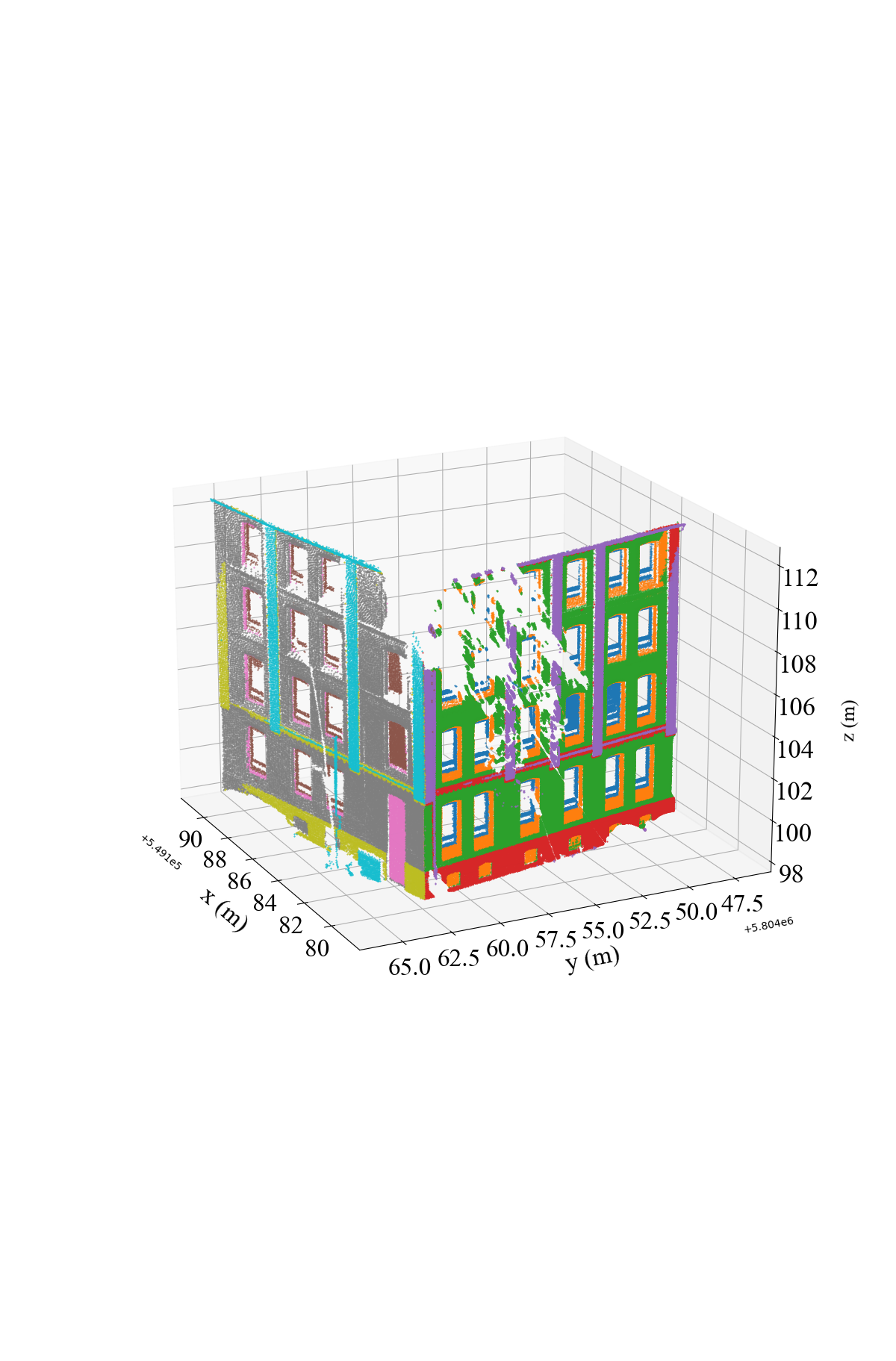}  
  \caption{Surface depth layers clustered by GMMs 
  ~(5 components for each façade, i.e., 5 layers)}
  \label{fig:lgmm}
\end{subfigure}
\begin{subfigure}[t]{.33\textwidth}
  \centering
  \includegraphics[width=0.73\linewidth, trim={3cm 12.8cm 0cm 15.5cm},clip]{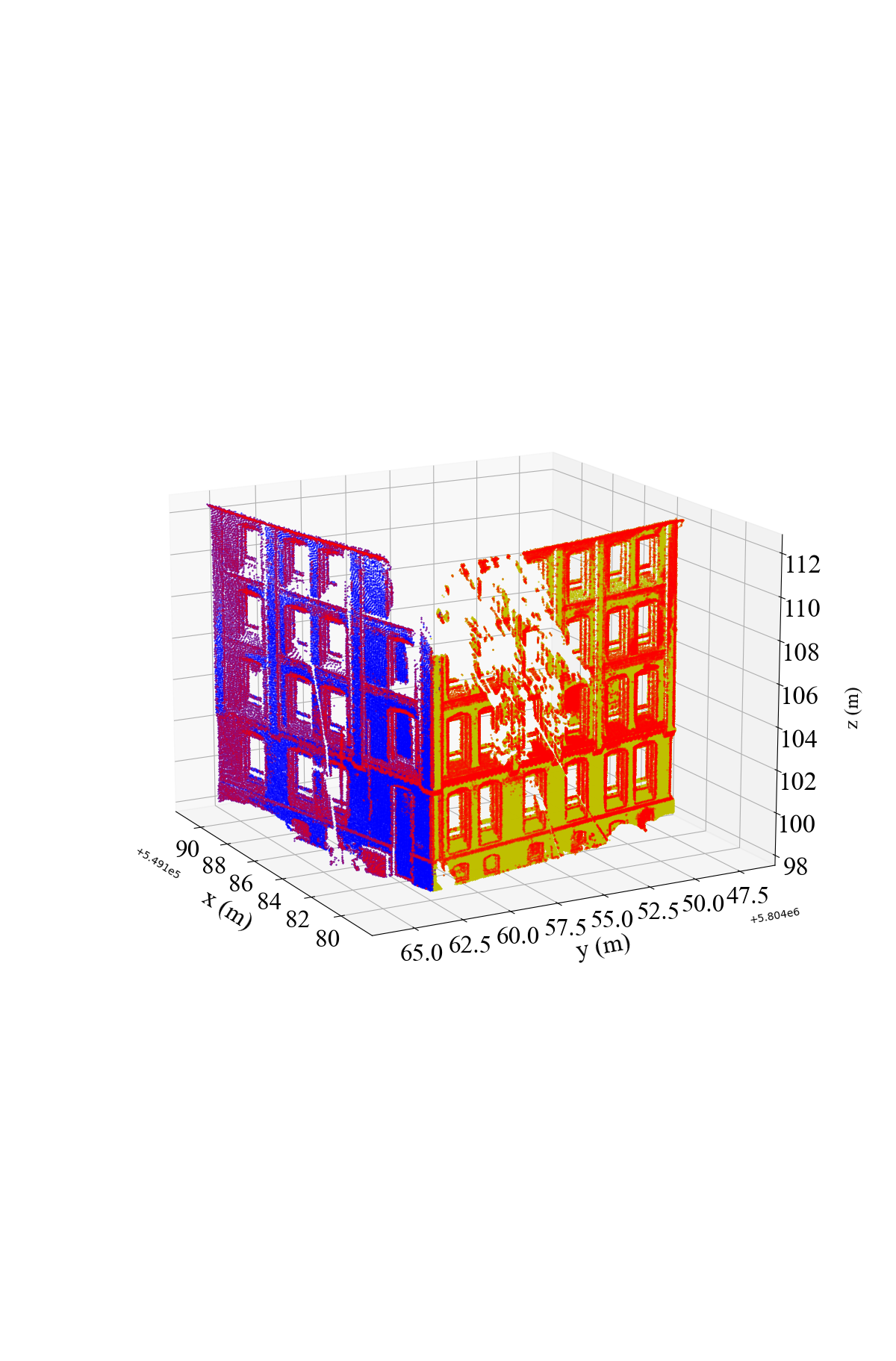}  
  \caption{Training points (red) in GPs}
  \label{fig:train}
\end{subfigure}
\begin{subfigure}[t]{.33\textwidth}
  \centering
  \includegraphics[width=0.73\linewidth, trim={3cm 12.8cm 0cm 15.5cm},clip]{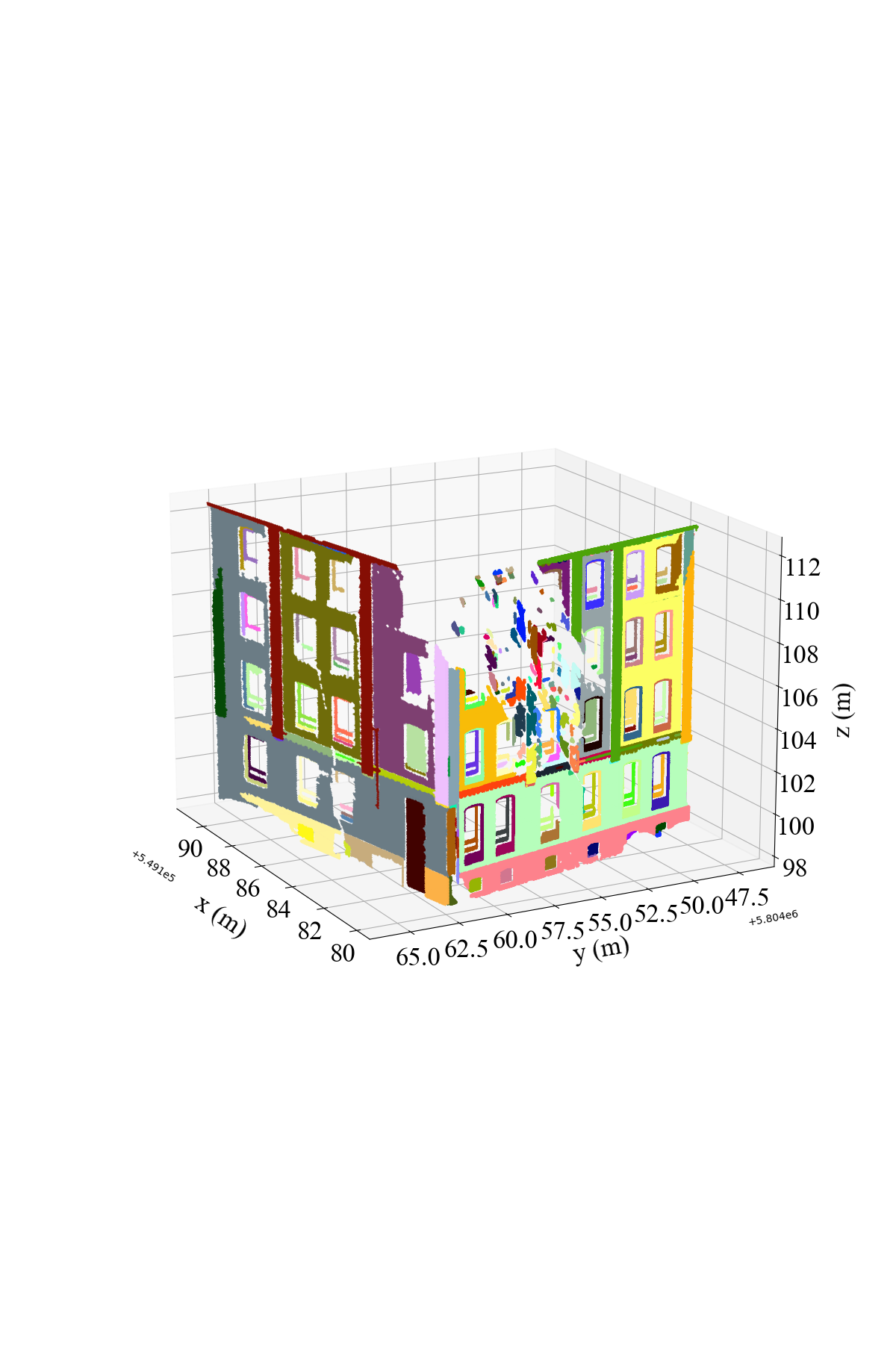}  
  \caption{Partitioning into local GPs}
  \label{fig:lgp}
\end{subfigure}
\begin{subfigure}[t]{.33\textwidth}
  \centering
  % include first image
  \includegraphics[width=0.73\linewidth, trim={3cm 12.8cm 0cm 14.5cm},clip]{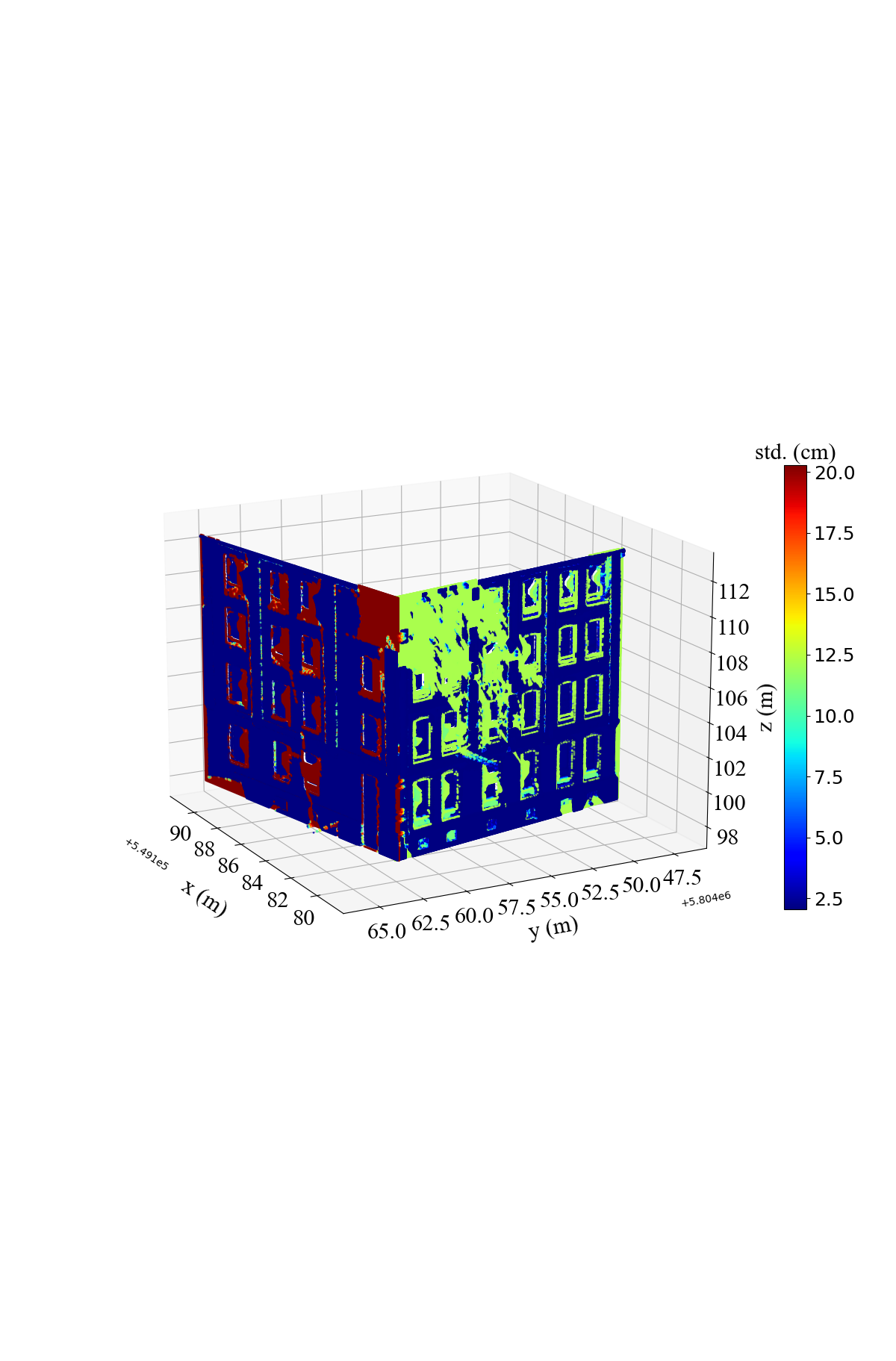}  
  \caption{Uncertainties of the estimated surface}
  \label{fig:result_s}
\end{subfigure}
% \begin{subfigure}[t]{.33\textwidth}
%   \centering
%   % include first image
%   \includegraphics[width=0.93\linewidth, trim={15cm 4cm 12cm 4cm},clip]{figures/results_m.png}  
%   \caption{Estimated surface: sparse or occluded areas in (a) are inferred with large uncertainty.}
%   \label{fig:result_m}
% \end{subfigure}
\begin{subfigure}[t]{.33\textwidth}
  \centering
  % include first image
  \includegraphics[width=0.73\linewidth, trim={3cm 12.8cm 0cm 14.5cm},clip]{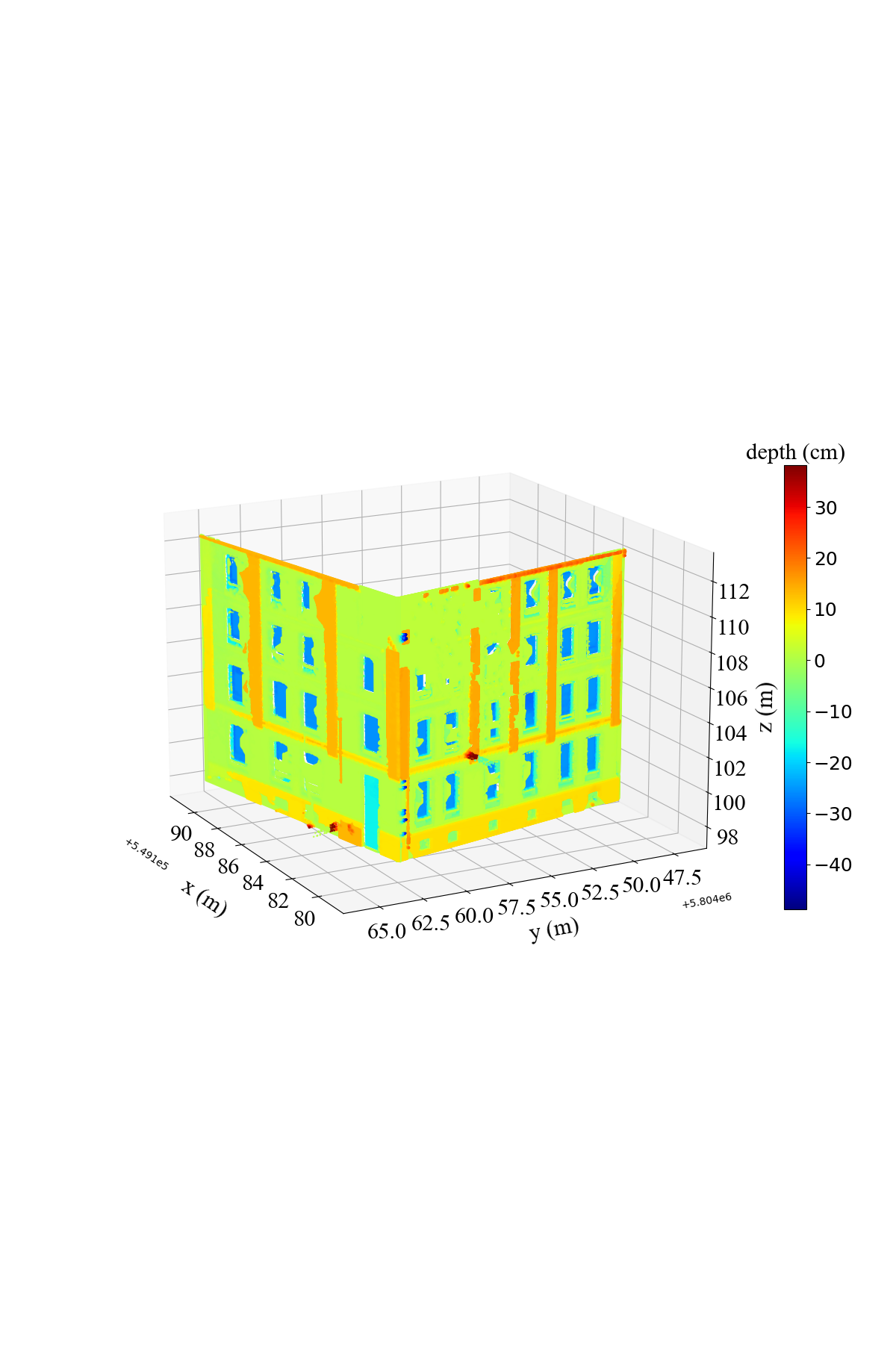}  
  \caption{Estimated surface: sparse or occluded areas in (a) are inferred with large uncertainty.}
  \label{fig:result_m}
\end{subfigure}
\begin{subfigure}[t]{.33\textwidth}
  \centering
  % include second image
  \includegraphics[width=0.73\linewidth, trim={3cm 12.8cm 0cm 14.5cm},clip]{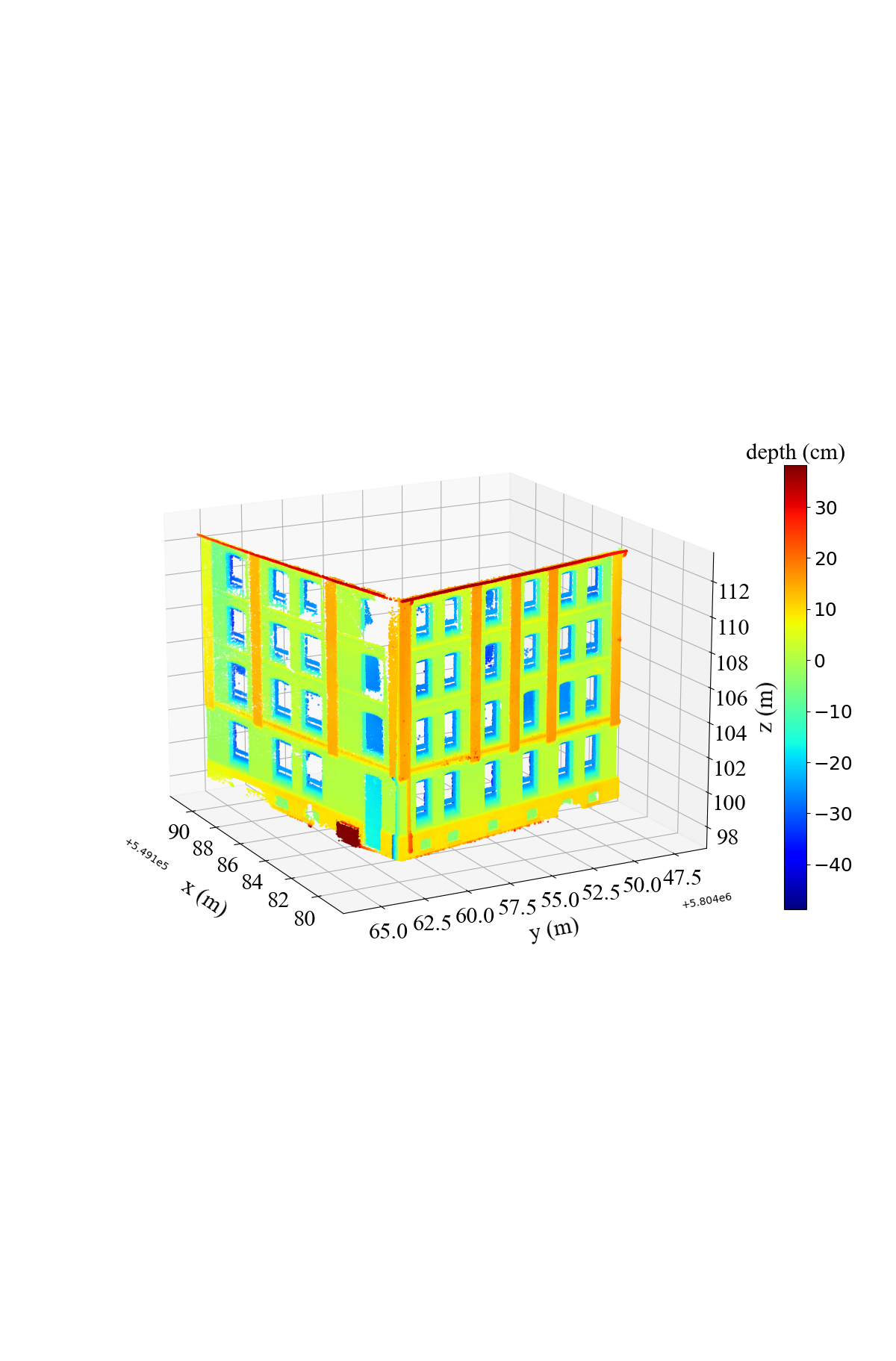}  
  \caption{Ground truth}
  \label{fig:gt}
\end{subfigure}
% \begin{subfigure}[t]{.33\textwidth}
%   \centering
%   % include first image
%   \includegraphics[width=0.93\linewidth, trim={15cm 4cm 12cm 4cm},clip]{figures/gt.png}  
%   \caption{Ground truth}
%   \label{fig:gt}
% \end{subfigure}
\caption{Visualization of different stages of the mapping process.}
\label{fig:workflow}
\end{figure*}

In addition to the depth value, the uncertainty is
%In the following, we illustrate how the uncertain building models are
represented probabilistically using GMM and GP components. Figure \ref{fig:workflow} visualizes the effect of each stage in the processing chain. To generate uncertain building models, the building points are first extracted by a semantic classification from calibrated MMS point clouds.  For each individual building, each façade plane is extracted by using a Random Sample Consensus (RANSAC) approach \cite{ransac}, where points are allocated to the corresponding façades. Then, an uncertain model of façades can be explored with GMMs and local GPs, illustrated in Figure \ref{fig:lgmm}, \ref{fig:train}, \ref{fig:lgp}, \ref{fig:result_s} and \ref{fig:result_m}. The ground truth of this exemplary building is provided for comparison, as shown in Figure \ref{fig:gt}.

\subsection{Gaussian Mixture Models for planar surfaces}

A façade is mainly composed of local planar objects, which can be modelled by the plane normal, location, and the boundary of the facets. The plane normal, i.e. the orientation of the building façade, is determined by a principal component analysis (PCA), using the eigenvector belonging to the smallest component.
It serves as the depth direction, which decides the local frame of the façade used in the following modelling. Many map representations such as 3D CityLoD2 models only map the building façade with one plane and ignore the extrusions and protrusions.
Often, the uncertainty is not specified at all, or if it is, is given in terms of a single, global uncertainty measure only, e.g. as used in \cite{interval_localization}.
This is not sufficient for many tasks requiring high accuracy. A GMM is applied to cluster the points with different depth values to depth layers. Each depth layer is described by one Gaussian component in the GMM:
\begin{equation}
    p(d)=\sum_{k=1}^{K}{\pi_k \mathcal{N}(d|\mu_k,\sigma^2_k )},
\end{equation}

where $d$ is the surface depth of a point, $K$ being the number of Gaussian components, which is determined by the number of peaks detected in the histogram of all depth values;
$(\pi_k, \mu_k,\sigma_k)$
are the GMM parameters for the $k$-th depth layer, $\pi_k$ is the prior probability or weight, representing the importance of the component, and $\mu_k$ and $\sigma_k$ are the mean and standard deviation of the Gaussian distribution.

\begin{figure}[ht!]
      \centering
      \includegraphics[width=0.58\linewidth, trim={1cm 0.2cm 0cm 0.7cm},clip]{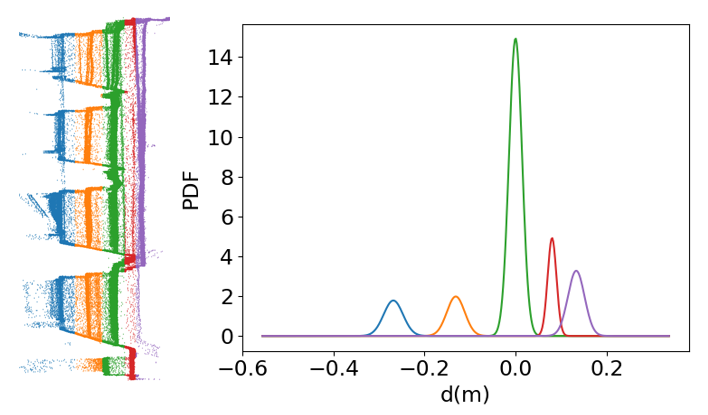}
      \caption{GMM for plane modelling: the left picture shows a façade
      profile, i.e., all scanned points of a façade in a view from the side;
      The Gaussian components (right) with different colors represent each depth layer. The one with highest peak is the main façade plane.}
      \label{fig:gmm}
 \end{figure}

The parameters
$\{(\pi_k, \mu_k,\sigma_k)\}_{k=1}^K$
are optimized by maximizing the log-likelihood using the EM algorithm. Figure \ref{fig:gmm} shows an example of the GMM results. The Gaussian component with the highest prior probability is the main façade plane.

Nevertheless, the limits of GMM are: 1)~Non-planar surfaces cannot be modelled. 2) If the number of points on different layers varies greatly, meaning the training for different components is not balanced, the method might fail to model the small protrusions and extrusions. 3) Slopes might be simply modelled as a vertical plane. The non-parametric approach GP, on the other hand, is capable to model them in a data-driven fashion. It is, therefore, introduced here to solve these problems.

\subsection{Gaussian Processes} 
A GP $f(x) \sim GP(\mu(x),k(x,x'))$ is a popular Bayesian nonlinear regression method in machine learning. It can be seen as a multivariate normal distribution with infinite variables in a continuous domain, e.g. a continuous space. The distribution of any finite subset of those variables is still a joint Gaussian distribution. It is characterized by a mean function $\mu(x)$ and a covariance function $k(x,x')$, which is also called the kernel function. The prior GP mean function is often set to zero. There are many possible choices for the kernel function
and in this letter, automatic relevance determination (ARD) is used as the prior GP kernel function:

\begin{equation}
\label{eq:kernel}
k_0(\boldsymbol{x},\boldsymbol{x}')= \sigma_p^2 \exp\left(-\frac{1}{2}\sum_{m=1}^{2}b_m(x_m-x_m')^{2}\right),	
\end{equation}
where
$\bm{\uptheta}=(\sigma_p,b_1,b_2)$
are the hyper-parameters for the kernel function, which are optimized by maximum likelihood, given a measured dataset.

Provided a measurement dataset
$\bm{D}=(\bf{X},\bm{y})$,
where ${\bf{X}}=\{{\bm{x}}_i\}_{i=1}^N$  are $N$ input vectors and  $\bm{y}=\{y_i \}_{i=1}^N$ are the corresponding target values measured with additive noise $\eta_i \sim \mathcal{N}(0,\sigma_\eta^2)$,
we have the following likelihood:

\begin{equation}
p(\bm{y}|{\bf{X}}, \bm{\uptheta})=\mathcal{N} (\bm{y}|{\bf K}_N+\sigma_\eta^2 {\bf I} ),
\end{equation}
where ${\bf K}_N= k_0 ({\bf{X}}, {\bf{X}})$ is the covariance matrix of the input points with $N\times N$ entries calculated by the prior kernel function.

The rest of the space $\bf{X}_*$ can be predicted with a multivariate Gaussian distribution conditioned on the observations $\bm{D}$. The posterior GP mean and kernel are calculated by \cite{gp-book}:

\begin{equation}
\mu(\boldsymbol{x}_*)=\mu_0(\boldsymbol{x}_*)+ {\bf k}_*^T ({\bf K}_N+\sigma_\eta^2 {\bf I} )^{-1} (\boldsymbol{ y}-\mu_0 ({\bf X})),
\end{equation}
\begin{equation}
k(\boldsymbol{x}_*,\boldsymbol{x}_*')=k_0 (\boldsymbol{x}_*,\boldsymbol{x}_*') - {\bf k}_*^T ({\bf K}_N+\sigma_\eta^2 {\bf I} )^{-1} {\bf k}_*'+\sigma_\eta^2,	
\end{equation}
where the prior mean $\mu_0 (\boldsymbol{x})$ is set to zero, $[{\bf k}_*]_N= k_0 ({\bf X}, \boldsymbol{x}_* )$ and  $[{\bf k}_*' ]_N= k_0 ({\bf X}, \boldsymbol{x}_*' )$ are the prior covariance between the $N$ input training points and the predicting points.

To utilize a GP for generating an uncertain building model from dense point clouds, there are mainly two problems. First, mapping methods using GPs have to consider the discontinuity of the environment while the GP often encodes the continuity and smoothness assumptions on targets. It is hard to model functions, which exhibit arbitrarily large derivatives, with a GP. Unfortunately, large derivatives often appear in environment mapping. This issue is also addressed in \cite{gpom, gpom_fast}, where they tried special kernels to adapt the discontinuities, e.g.\ Mat\'ern covariance functions. However, the improvement is limited. Second, with large training datasets, the GP regression faces a computational problem due to the inversion matrix calculation having time complexity $O(N^3)$. To capture the discontinuity and reduce the computational cost of GPs in the case of a building model, the two strategies applied here are: 1)~using the GMM results as prior, and 2)~partitioning data into blocks. These strategies are explained in detail in the following sections.

\subsection{Using the GMM as prior}

% Since buildings have relatively simple shapes with many planar objects, these planar parts can be modelled by a GMM as introduced in Subsection III.A. 
Using the GMM as prior offers several advantages. The GMM is used to “explain” the planar building parts. Thus, only the remaining arbitrary surfaces have to be modelled by GP, namely non-planar points, points of non-horizontal normals and outlier points too far away from the mean. In this way, the GMM  information reduces the number of training points for GPs, as only the irregular points are used for GP training. 

Another advantage of using GMM as prior is to provide parameters to the prior distributions of GPs. The estimated $\mu$-values from the GMM serve as prior means in GP modelling, denoting the prior surface depth expectation of each planar layer, while the variances parameterise the prior covariance functions and depict the local data variability. 
%Additionally, the planar segments with sharp depth variance compared to the neighboring regions cause discontinuities on building surfaces. %Another advantage of using GMM results as prior is to capture and parameterise the discontinuities on the façade surface. The independence can be introduced in different mean and covariance functions that are applied in different areas of the input locations. 
The non-planar points which fall into a local facet on a certain layer are modelled by a local GP with this prior mean. The $g(\boldsymbol{x})$ in Equation \eqref{eq:statement} is substituted by the mean $\mu_k$ of a GMM component. Each local GP model has its own hyper-parameters and estimates the predictions only for the corresponding area. All local GPs compose a GP list for inference.

Different mean and covariance functions adapt GPs to the local facets, capturing and parameterising the discontinuities on the facade surface, caused by the sharp change between two depth layers. With this process, the entire continuous space is also divided into multiple local areas of GMM facets, modelled by GPs individually, which reduces the size of the kernel matrix as well. 

There might be cases, where there are regions without any training data and not covered by any local GPs. Then, a global GP prior is included in our GP list. The mean value of this GP is determined by the mean of the Gaussian component with the largest weight in the GMM. Its prior variance can be set as the variance of the entire façade plane, representing the overall variability across all layers of the facade.

Initially, we considered weighted averaging of the inferred results from adjacent local GP segments in empty areas, using either BCM \cite{grbcm} or Product of Expert (PoE) \cite{poe}. However, in our case, the fusion did not significantly improve the results, and the resulting decrease in uncertainty was not desirable. Furthermore, the use of mixtures of GPs would have impacted efficiency. Therefore, in this letter, mixtures of local GPs are not utilized.

We classify the points that cannot be effectively modelled by the GMM into three categories: (1)~non-planar points, (2)~non-vertical slopes whose normal vectors deviates considerably from the normal of the main plane, and (3)~points that are significantly outside their Gaussian distributions, i.e. not within the 95\% confidence interval ($1.96\sigma_k$) of a certain component. Figure \ref{fig:train} illustrates an example, where the red points are selected as training points and model the GP implicit surface as a function of the continuous surface depth estimate. These points are selected according to the rules regarding the depth value $d$ and the normal vector ${\bf{n}}=\{n_x,n_y,n_z\}$:
\begin{equation}
\label{eq:condition1}
|d-\mu_k |>1.96 \sigma_k,
\end{equation}

\begin{equation}
\label{eq:condition2}
n_z ({\bm{x}})< \cos (\alpha_z),  
\end{equation}
where $\mu_k$ and $\sigma_k$ are the parameters of the $k$-th Gaussian layer, and $n_z ({\bm{x}})$ is the component of the normal $\bf{n}$, which is perpendicular to the main façade plane. $\alpha_z$ is the angle threshold for the normal ${\bf{n}}$ deviating from the normal of the assigned GMM plane. In this letter, $\alpha_z = 25^{\circ}$ is applied.
%If the normal vectors of training points deviate with an angle larger than $\alpha_d$, the points will be seen as training points.

The estimate for $\sigma_p$ in the kernel of a local GP is set as the standard deviation of the corresponding planar patch $\sigma_k$. Optimization of other hyper-parameters is done on a small subset of the training data, involving a reasonable choice of the initial estimates:

\begin{itemize}

\item The initial $\{b_1,b_2\}$ are set to the same value, i.e. isotropic. We set $\frac{1}{b_1}=\frac{1}{b_2}=0.0025\,m^{2}$, observed from the investigation of the impact of the length-scale $l^{2}=\frac{1}{b_1}=\frac{1}{b_2}$ on kernels and the empirical value from the existing study \cite{gpom}; 
\item $\sigma_\eta$ reflects the precision of the sensors, point cloud alignment noise and random environmental noise. In this case, $\sigma_\eta = 0.02\,m$. 

\end{itemize}

\subsection{Applying GP-block techniques}

Although we reduced the training points in the last section, the computational efficiency can be further improved by the data partitioning and GP-block techniques, which are inspired by \cite{gpmap}. In practice, the spatial correlation between two points decreases with their distance. This property is naturally captured by the length-scale hyper-parameter
$\bm{b} = (b_1, b_2)$
in the covariance functions of the GPs. As the points far from each other have negligible influence that can be ignored in the covariance matrix, only close points have significant covariance and need to be considered. An idea to exploit this property is to apply a sparse covariance function \cite{sparsekernel}, by setting the covariance to zero when the distance is larger than a threshold. It also indicates the possible data partition of the training data. The entire training samples can be divided into local blocks within a certain range, as proposed in\cite{gpmap,gpom_fast}.  In those methods, the block size is set as the predefined scale value of the sparse kernel while in our case, the block size is determined by the distance corresponding to the threshold $c_{min}$ of covariance values, calculated by Equation \eqref{eq:kernel}, when $\sigma_p$ is set to 1 and $b_1=b_2=b$:
\begin{equation}
\exp\left(-\frac{1}{2}\sum_{m=1}^{2}b(x_m-x_m')^{2}\right) \geq c_{min},	
\end{equation}
where $c_{min}=0.0001$ in this case.
The block size is equal to $|\bm{x}-\bm{x}'|$ when it makes both sides equal.
The extended block of a given block is defined as the central block with its neighboring blocks. In GP regression, the points from the extended block are the training data for the central block being under consideration.

\subsection{Outlier filtering}
To generate a GP inference robust to outliers, a model is required for outlier processing. There might be outliers in the training samples due to false classifications (e.g. points of road signs in the vicinity of a building) or other unexpected reasons. A statistical test on a GP with Gaussian likelihood is applied to check the probabilistic significance and identify the outliers, e.g. a Chi-squared test. In this work, the Chi-squared test with one degree of freedom is applied to remove outliers.

The procedure of outlier filtering is as follows.
\begin{enumerate}
\item The GP is trained normally with all training data and the mean and variance of each training point are inferred,
\item The Chi-squared test-statistic is calculated
using $v_i^{2} = (d_i - \mu_i)^{2} / \sigma_i^{2}$,
\item The test-statistics are compared to the $\chi^{2}$ value given a certain $\alpha$-quantile level and the points with test-statistic values larger than $\chi^{2}$ value will be removed as outliers,
\item The GP is trained again with the filtered training data and the test points are predicted with mean and variance.
\end{enumerate}

The above process may as well be repeated to iteratively trim the outliers,
which, however, will lead to an increased computation time.
Therefore, in this case, we only perform the test once in our experiment. The $\alpha$-percentile level in the test serves as a threshold to filter outliers, indicating the strictness of outlier detection. High values of $\alpha$ will see many points as outliers, while a small $\alpha$ gives more tolerance to the points deviating from the mean and might miss the outliers.

\section{Experiments and Evaluation}
\subsection{Dataset}
The GP-based representation of a building surface is evaluated using real-world LiDAR point clouds collected by a Riegl VMX-250 mobile mapping system for the urban environment, in Hanover, Germany. The scanner measurements are specified as $5\,mm$ precision and $10\,mm$ accuracy. However, the accuracy of the GPS sensor in the system is much worse than the laser scanners. Point cloud alignment is thus used to correct the GPS trajectories in preprocessing and the data noise is given by a $2\,cm$ standard deviation, resulting from the alignment \cite{alignment}. In the experiments, only building points were used, which were extracted using the deep learning based method \cite{classification}. 

In total, we used five measurement campaigns for the same urban areas collected on different dates to obtain very dense point clouds. 2,239,670 points from one measurement campaign were used for modelling the uncertain map and the remaining data (11,694,768 points) was used as ground truth for testing and evaluation. In one single measurement campaign, the number of points for one building façade is often larger than 10,000 and can reach up to 450,000 in our dataset. The investigated urban area is around $55\times93\times22\,m^{3}$.

\subsection{Experiments}
 
 The building façades obtained from the RANSAC, are firstly transformed individually to the local frame of their main planes, where a map point is a 2.5-dimensional point $\mathcal{P}=\{\bm{x} ,d\}$, with the location $\bm{x} \in \mathbb{R}^2$ on the plane and the depth value $d$ in the direction of the plane normal (derived by the PCA using all façade points).

In the local façade frame, the planar points are used as i.i.d. samples to optimize the parameters of the GMM. Training points selected according to Equation \eqref{eq:condition1} and \eqref{eq:condition2} are used to derive the posterior GP-based models. For different building façades, the amount of training points varies. Depending on the building structures, $20$ to $50\%$ of the points are selected as training points in the experiment. The initial hyper-parameters are given based on the rules in Section~III.C.  The initial $\sigma_\eta$ is chosen as $2\,cm$ for this dataset.

\begin{figure}[ht!]
    %   \centering
    %   \includegraphics[width=0.48\linewidth]{figures/qualitative.png}
    % %   \includegraphics[scale=1.0]{figures/GMM.png}

\begin{subfigure}{.241\textwidth}
  \centering
  % include first image
  \includegraphics[width=0.95\linewidth]{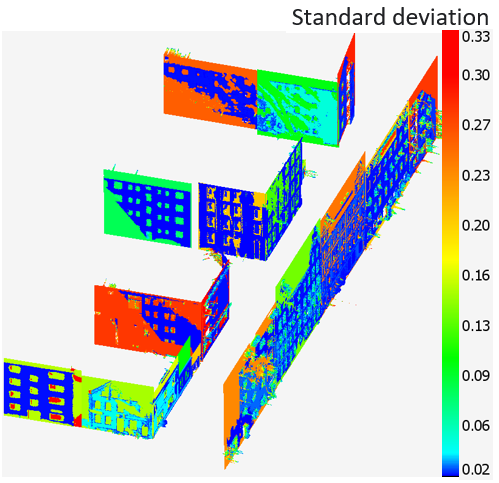}
  \caption{Surface inference}
  \label{fig:qualitative1}
\end{subfigure}
\begin{subfigure}{.241\textwidth}
  \centering
  % include second image
  \includegraphics[width=0.95\linewidth]{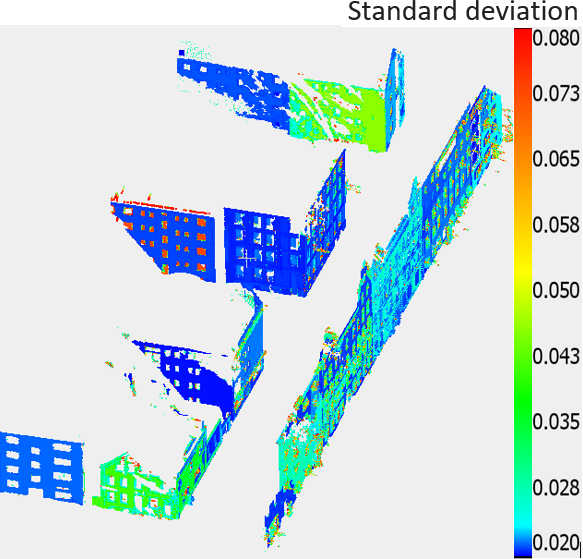}
  \caption{Accurate inference}
   \label{fig:qualitative2}
\end{subfigure}
\caption{Qualitative results of the building models: uncertainties are denoted by colors. Redder colors indicate higher uncertainties while blue colors denote low uncertainties.}
   \label{fig:qualitative}
 \end{figure}

Figure \ref{fig:qualitative} shows the qualitative result of modelled buildings with their associated uncertainties. In the original measurements, certain areas are occluded, where the inferred surfaces exhibit high uncertainties. Among the occluded regions, those located on simple façades with smaller prior variances, have lower uncertainties, since the variance of the entire façade is indicative of the potential surface variability, and hence, the possible errors. In general, the buildings that have been observed with denser measurements are modelled with lower uncertainties. In practical localization applications, users are able to select only the accurate part of maps or use the entire reference map with weights indicating the uncertainty. The benefits of only using maps with low uncertainties in occupancy maps are investigated in the following evaluation section.

\subsection{Evaluation}
To evaluate the proposed method, existing state-of-the-art approaches are used as benchmarks, namely OctoMap \cite{octomap}, GPOM \cite{gpom_fast}, BGKOctoMap \cite{bgk}, LARD-HM \cite{hm1} and GPIS map \cite{gpis1}. To make the comparison possible, we first transform our abstract surface map to a 3D occupancy map, where probabilistic least-squares classification is applied and the occupancy probabilities are obtained through a cumulative Gaussian density function, inspired by previous research \cite{gpom,gpmap}. The details can be found in the paper \cite{gpmap}.

The metrics used for evaluation are 1)~Precision-Recall (PR) curves and 2)~Area under the curve (AUC). Note that some previous work used the ROC curve and the corresponding AUC as the metrics. However, it is not proper to use ROC in highly-skewed data, where a large change in the number of false positives can only lead to a small change in the false positive rate used in ROC \cite{rocpr}. This occurs in our case, as in outdoor scenarios, the number of free samples is much larger than the number of occupied ones. This issue has also been addressed in previous research \cite{metric}. Alternatively, the PR-curve is more suitable for highly imbalanced data \cite{rocpr}, especially when the positive (occupied) case is of more concern. Precision compares the false positives with true positives, and therefore, captures the detection errors of a large number of false positives. 
Unlike the case with ROC curves, the goal of PR curves is to reach the upper right corner, instead of the upper left corner.

\begin{figure}[ht!]
\begin{subfigure}{.241\textwidth}
  \centering
  % include first image
  \includegraphics[width=.93\linewidth, trim={0.cm 0.cm 0.8cm 1.2cm},clip]{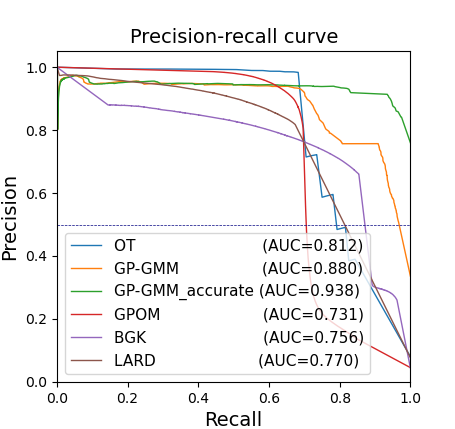}  
  \caption{Resolution: 0.1m}
  \label{fig:pr0.1}
\end{subfigure}
\begin{subfigure}{.241\textwidth}
  \centering
  % include second image
  \includegraphics[width=.935\linewidth, trim={0.cm 0.cm 0.8cm 1.2cm},clip]{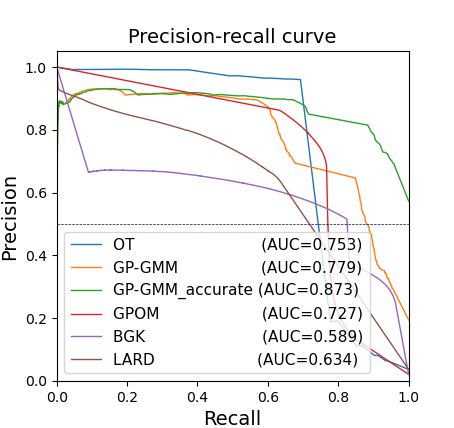}  
  \caption{Resolution: 0.05m}
  \label{fig:pr0.05}
\end{subfigure}

\caption{PR curves and AUC comparison: the AUC value of each curve is given in the legend.}
\label{fig:pr_curve}
\end{figure}

Figure \ref{fig:pr_curve} compares the AUC of different approaches in two resolutions ($0.1\,m$ and $0.05\,m$ grids). In general, GP-based maps have a good performance, among which our proposed method (green and orange curves) achieves the highest AUC values and is robust to small resolutions. The green curves denote the case when we only consider the accurate inference, i.e. areas with small uncertainty. The variance threshold is selected as 0.01 in this case. Areas with inference variance larger than 0.01 are considered being unknown. The figure shows that improvement is obtained by using only the accurate parts and the corresponding maps (green curves) give the best performance. This shows the value of the uncertainty representation for maps and indicates the potential application of this uncertainty model in downstream uncertainty-aware tasks.

%***********************revision*********************

\begin{figure}[ht!]
\begin{subfigure}{.241\textwidth}
  \centering
  % include first image
  \includegraphics[width=.88\linewidth, trim={0.21cm 0.cm 1.0cm 0.9cm},clip]{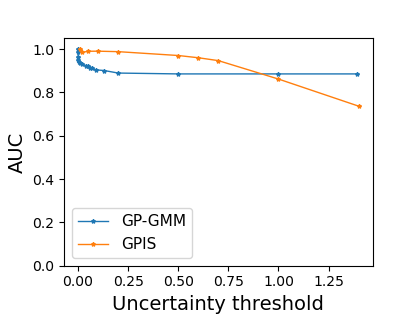}  
  \caption{}
  \label{fig:auc_u}
\end{subfigure}
\begin{subfigure}{.241\textwidth}
  \centering
  % include second image
  \includegraphics[width=.88\linewidth, trim={0.21cm 0.cm 1.0cm 0.8cm},clip]{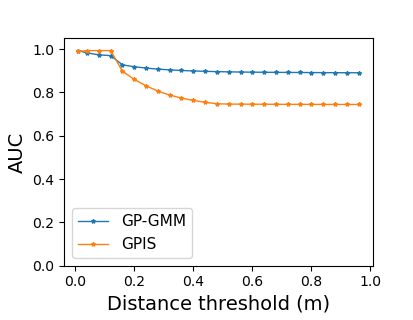}  
  \caption{}
  \label{fig:auc_n}
\end{subfigure}

\caption{Uncertainty evaluation: change of AUC with varying uncertainty thresholds and distances between the predicted points and original training points.}
\label{fig:auc_compare}
\end{figure}

The uncertainty evaluation is conducted by plotting AUC changes across varying uncertainty thresholds and distances of query points from the nearest observations. The proposed work is compared with the benchmark - GPIS map, shown in Figure \ref{fig:auc_compare}. The results reveal a clear monotonic trend in the AUC values as the uncertainty threshold varies. The proposed method shows robustness compared to GPIS for the areas far from the observations, where the uncertainty is high. The AUC values remain consistently high, indicating strong performance. In contrast, GPIS exhibits a higher AUC only when a small number of predicted points, which are close to the original training points, are retained, as shown in Figure \ref{fig:auc_n}. However, our method maintains a high AUC even with a larger number of predicted points, suggesting its ability to deliver reliable predictions for unobserved areas around observation points.

\subsection{Comprehensive comparison}
The GPOM and BGK models use fixed values for prior variance, noise, or length scale, which may not adequately support accurate spatial correlation and data variability. Additionally, the fixed prior variance and kernels struggle to model sharp changes in voxel occupancy, potentially impacting the final posterior occupancy probability. Similarly, the GPIS benchmark sets pre-fixed values for prior variance and mean, introducing inaccuracies in surface estimation, particularly for regions far from the observations. The fixed variance also leads to potential overestimation or underestimation of uncertainties. Thus, the results are sensitive to these fixed priors.  

By contrast, our proposed method addresses these limitations by leveraging the GMM for obtaining a more tailored prior variance and mean, and employing local GPs for capturing the spatial correlations and uncertainties in a more adaptive manner. This allows for a more accurate estimation of façade surfaces.

\subsection{Computational time comparison}

\begin{figure}[ht!]
\begin{subfigure}{.241\textwidth}
  \centering
  % include first image
  \includegraphics[width=.93\linewidth, trim={0.0cm 0.cm 1.0cm 0.9cm},clip]{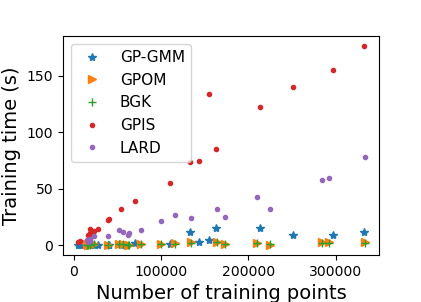}  
  \caption{Training time}
  \label{fig:train1} % fig:train already used above
\end{subfigure}
\begin{subfigure}{.241\textwidth}
  \centering
  % include second image
  \includegraphics[width=.93\linewidth, trim={0.1cm 0.cm 1.0cm 0.9cm},clip]{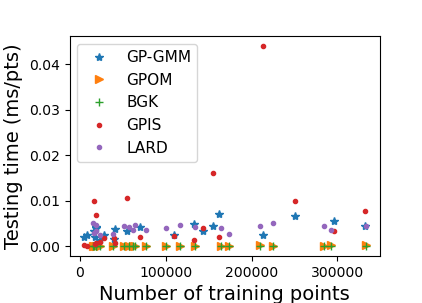}  
  \caption{Testing time}
  \label{fig:test}
\end{subfigure}

\caption{Computational time.}
\label{fig:time_compare}
\end{figure}

Regarding computational time, BGK and GPOM reach the fastest speed for mapping, as shown in Figure \ref{fig:time_compare}. Our proposed method demonstrates comparable training times to benchmarks, thanks to the assistance of the GMM prior. During testing, it may require more time than BGK, GPOM, and LARD-HM but remains faster than GPIS. While we acknowledge the importance of efficient online mapping, our goal is to deliver a mapping framework that prioritizes accuracy and uncertainty awareness, enhancing map-based localization tasks. 

\section{Conclusions}

In summary, this letter proposed a method to generate building models as reference maps with associated uncertainties. The GMM and GP approaches are combined to map the planar parts and non-planar surfaces together. Local GP and data partition techniques are introduced in GP to reduce the computational cost and parameterise the discontinuity of the building surface. In the experiments, the abstract surface map is transformed into an occupancy map representation and then is compared with state-of-the-art benchmarks. Our evaluation shows that the local GP with GMM prior yields the highest AUC values, indicating its potential for utilizing uncertainty characterizations in future applications. 

The generated map will be applied for localization tasks in the future and the positioning results will be compared with other methods, such as NDT-based localization. The approach also has a large potential to be used for updating and refining existing models from new data or other sensors; this will also be studied in future work.

%%%%%%%%%%%%%%%%%%%%%%%%%%%%%%%%%%%%%%%%%%%%%%%%%%%%%%%%%%%%%%%%%%%%%%%%%%%%%%%%

\section*{Acknowledgments}

The authors acknowledge Jinkun Wang, Kevin Doherty, Erik Pearson, Bhoram Lee and Vitor Guizilini for making their code available and kindly clarifying their methods.

% \newpage

\vfill

\end{document}